\documentclass[letterpaper, 10 pt, conference]{ieeeconf}  % Comment this line out if you need a4paper

\IEEEoverridecommandlockouts                              % This command is only needed if 
\usepackage{cite}
\usepackage{amsthm} 
\usepackage{amsmath,amssymb,amsfonts}
\usepackage{graphicx}
\usepackage{textcomp}
\usepackage{xcolor}

\usepackage{algorithm}
\usepackage{algpseudocode}

\usepackage{todonotes}

\usepackage[font=small,skip=0pt]{caption}
\usepackage{subcaption}
\theoremstyle{definition}
\newtheorem{definition}{Definition}
    
\algnewcommand\algorithmicforeach{\textbf{for each}}
\algdef{S}[FOR]{ForEach}[1]{\algorithmicforeach\ #1\ \algorithmicdo}

\DeclareMathOperator{\E}{\mathbb{E}}

\title{\LARGE \bf Learning Stabilization Control from Observations by Learning Lyapunov-like Proxy Models}

\author{Milan Ganai, Chiaki Hirayama, Ya-Chien Chang, and Sicun Gao%
\thanks{This material is based on work supported by DARPA Contract No. FA8750-18-C-0092, AFOSR YIP FA9550-19-1-0041, NSF Career CCF 2047034, NSF CCF DASS 2217723, and Amazon Research Award.}% <-this % stops a space
\thanks{M. Ganai, C. Hirayama, Y.-C. Chang, and S. Gao are with the Department of Computer Science and Engineering, UC San Diego, USA. (Email: mganai@ucsd.edu; chirayam@eng.ucsd.edu; yac021@eng.ucsd.edu; and sicung@eng.ucsd.edu).}%
}

\begin{document}

\maketitle
\thispagestyle{empty}
\pagestyle{empty}

%%%%%%%%%%%%%%%%%%%%%%%%%%%%%%%%%%%%%%%%%%%%%%%%%%%%%%%%%%%%%%%%%%%%%%%%%%%%%%%%
\begin{abstract}

The deployment of Reinforcement Learning to robotics applications faces the difficulty of reward engineering. Therefore, approaches have focused on creating reward functions by Learning from Observations (LfO) which is the task of learning policies from expert trajectories that only contain state sequences. We propose new methods for LfO for the important class of continuous control problems of learning to stabilize, by introducing intermediate proxy models acting as reward functions between the expert and the agent policy based on Lyapunov stability theory. Our LfO training process consists of two steps. The first step attempts to learn a Lyapunov-like landscape proxy model from expert state sequences without access to any kinematics model, and the second step uses the learned landscape model to guide in training the learner's policy. We formulate novel learning objectives for the two steps that are important for overall training success. We evaluate our methods in real automobile robot environments and other simulated stabilization control problems in model-free settings, like Quadrotor control and maintaining upright positions of Hopper in MuJoCo. We compare with state-of-the-art approaches and show the proposed methods can learn efficiently with less expert observations.

\end{abstract}

%%%%%%%%%%%%%%%%%%%%%%%%%%%%%%%%%%%%%%%%%%%%%%%%%%%%%%%%%%%%%%%%%%%%%%%%%%%%%%%%

\section{Introduction}

%%
%% Modified New intro
%%
Reward engineering remains a challenge in applying Reinforcement Learning to robotic control problems~\cite{dewey2014reinforcement}. Manual design of suitable reward functions can be complicated, requiring extra sensors~\cite{Yahya2017, Kormushev2010RobotMS} and may guide learner to acquire unintended behavior~\cite{amodei2016concrete}. Consequently, imitation learning has become an indispensable approach that allows usage of expert demonstrations to replace reward functions, typically in the setting of learning from demonstrations (LfD)~\cite{Argall2009ASO, ILSurvey}, where the learner has access to several state-action pair sequences of desired behavior generated by the expert. However, actions may be difficult to retrieve (like with human experts) and there can be mismatch between expert and learner action spaces. Therefore, methods have been proposed for the problem of learning from observations (LfO) where access to expert behavior is limited to state sequences~\cite{RecentAdvILfO}. LfO can model many challenging forms of imitation such as learning from videos to acquire control skills. State-of-the-art methods for LfO typically adapt methods from LfD to the state observation setting. For instance, the Generative Adversarial Imitation from Observation algorithm (GAIfO)~\cite{RecentAdvILfO, GAIfO} adapts Generative Adversarial Imitation Learning (GAIL)~\cite{GAIL} by using Generative Adversarial Networks (GANs)~\cite{GAN} to learn policies that mimic expert state transition distributions. However, because of the absence of actions, models, and reward signals, LfO problems are often too challenging in the general continuous control settings~\cite{GAIfO}. Our goal is to develop efficient LfO methods for a specific but important class of continuous control problems: stabilization control, such as controlling a robot to maintain an upright standing pose, or controlling an autonomous car to track a given path. Stabilization is the basis of all advanced control problems~\cite{lyapunovbook}. Complex tasks can almost always be decomposed into a motion planning part and a stabilization control part, which is used for following the plans.
\begin{figure}[t!]
\centering
\includegraphics[width=\linewidth]{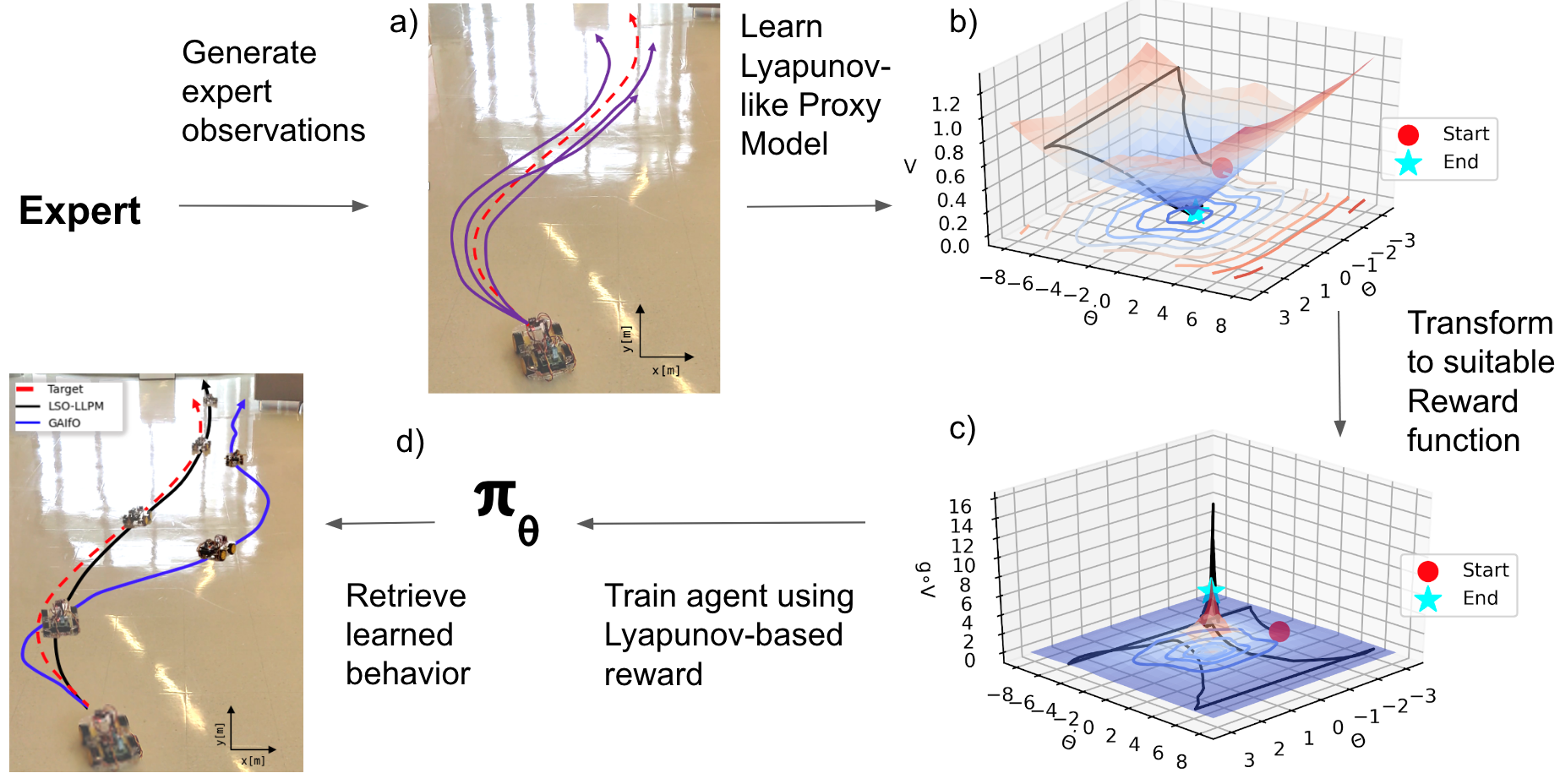}
\caption{ (a) We collect some expert state trajectories. (b) We learn a Lyapunov-like proxy model from the state trajectories. (c) We transform the model to a reward function. (d) We subsequently train the agent to learn a policy using the reward function. In the bottom left image, we show a trajectory from proposed approach compared with that of state-of-the-art GAIfO~\cite{GAIfO}.}

\label{fig:lyaplandscape}
\end{figure}

We propose a novel two-step LfO procedure for stabilization control to achieve efficient imitation learning, supported by control-theoretic principles. We exploit the framework of Lyapunov stability theory, which provides a general structure for stabilization in nonlinear control systems~\cite{lyapunovbook}. The core idea of Lyapunov methods is to construct an energy-like landscape that provides sufficient conditions for control systems to converge to and stabilize at their equilibrium points. In the LfO setting, we propose methods that learn Lyapunov-like functions as an intermediate goal, which we name as {\em Lyapunov-like proxy models}. From observations of expert state sequences (Figure~\ref{fig:lyaplandscape} (a)), we first learn a neural network model that attempts to construct an approximate Lyapunov landscape to explain convergence of expert states (Figure~\ref{fig:lyaplandscape} (b)). We transform this learned landscape model (Figure~\ref{fig:lyaplandscape} (c)) to guide the training of the learner policy (Figure~\ref{fig:lyaplandscape} (d)). For this procedure to succeed, it is critical for the proxy model to capture the geometry of the region of attraction and convergence rates (i.e. Lie derivatives of Lyapunov function) so they are consistent with expert behavior dynamics. We test our proposed procedure in real and simulated environments.

In the context of GAN-based approaches for LfO and LfD~\cite{GAIL, Fu2018, Henderson2018, Kostrikov2019, Sun2021}, we can consider the Lyapunov-like proxy models as introducing a special hypothesis class for the discriminators, as opposed to using generic distributions, to achieve more robust modeling of the expert behaviors for the specific context of stabilization problems.

We describe our technical approach in Section~\ref{section:AlgExplanation}. In Section~\ref{section:Experiments}, we evaluate our approach in challenging environments for learning to stabilize from observations, including Acrobot, Quadrotor control, Automobile path-tracking, and stabilization version of MuJoCo Hopper 
%and Walker
robot. We evaluate our approach on real automobile robot environments in Section~\ref{section:ExperimentsHW}. Compared to state-of-the-art methods, the proposed methods can learn more efficiently from less observations of expert trajectories and produce more stable control policies.

\section{Related Work}
\label{section:RelatedWork}

\noindent{\bf Imitation Learning.} Learning from Demonstrations (LfD) problems have been the most well-studied form of imitation learning in MDPs. The learner has access to state-action trajectories of the expert without knowledge of the transition dynamics or the reward function in the MDP. Existing approaches in LfD generally fall into three categories: Behavioral Cloning~\cite{Bain1995, Pomerleau1988}, Inverse Reinforcement Learning~\cite{Abbeel2004, Ng2000}, and Adversarial Imitation Learning~\cite{GAIL, Fu2018, Henderson2018, Kostrikov2019, Sun2021}. Our approaches are most related to the last category, where imitation learning is formulated in an adversarial framework of learning the policy as a generative model from the simultaneous training of a generator model and a discriminator model. The state-of-the-art algorithms are typically based on the method of Generative Adversarial Imitation Learning (GAIL)~\cite{GAIL}, which uses Generative Adversarial Networks (GANs)~\cite{GAN} to train a generative model that can create trajectories with a state-action occupancy measure similar to that of the expert, while the discriminator learns to provide feedback signals by differentiating the behavior distributions between the expert and the learner. Discriminator Actor-Critic (DAC)~\cite{DAC} is an off-policy version of GAIL that uses state-transition samples to train off-policy to achieve mode-covering in distribution matching. In general, LfD does not handle the most challenging forms of imitation, like learning from visual data of only expert state observations.

\noindent{\bf Learning from Observations (LfO).} GAN-based approaches from LfD can be extended to the LfO setting since such methods can be used to only imitate state distributions without access to actions from the expert. The Generative Adversarial Imitation from Observation (GAIfO) algorithm~\cite{RecentAdvILfO, GAIfO} uses a similar GAN framework as GAIL. Instead of training the discriminator using state-action pairs, GAIfO uses state transitions so that the imitator policy, which is the generator, produces a similar distribution of state transitions as expert.  Another approach to LfO is Behavior Cloning from Observation (BCO)~\cite{BCO} where the agent acquires inverse dynamics experience in a self-supervised manner, which is then used to create a model to perform a task based on expert state observations. This approach does not require post-demonstration environment interactions, so it reduces the delay before the imitating agent is successful and avoids training and learning in risky environments (like autonomous vehicles). BCO nevertheless faces the issue of inaccuracies in that the learned inverse dynamics model accumulates error over time~\cite{Laskey2016,RossBagnell2010,Ross2011}. GAIfO has been shown to perform better than BCO by alleviating this compounding error issue~\cite{RecentAdvILfO, GAIfO}.
Off-policy learning methods have also been introduced to the GAN-based approaches for LfO. The works of Off-Policy Imitation Learning from Observations (OPOLO)~\cite{OPOLO} and Inverse Dynamics Disagreement Minimization (IDDM)~\cite{IDDM} are able to accelerate the training of the learner/generator in the GAN framework by leveraging off-policy training of the inverse dynamics or action models of the environment before imitating from the expert. The works of Forward Adversarial Imitation Learning (FAIL)~\cite{FAIL} and Imitating Latent Policies from Observation (ILPO)~\cite{ILPO} also learn forward dynamics models and assume environment dynamics is deterministic with discrete action spaces. In contrast to this line of work, we focus on efficient learning in the setting of on-policy LfO without learning models or leveraging off-policy samples. The performance gain from off-policy methods can be used orthogonal to ours. 
 
\noindent{\bf Lyapunov-based Methods in Reinforcement Learning and Imitation Learning.} 
%\todo{to double check and see if all the recent approaches have been mentioned} 
Lyapunov-based approaches have been recently introduced in model-free learning tasks~\cite{Chow2018, DBLP:journals/corr/abs-1901-10031, DBLP:journals/corr/abs-2004-14288, jin2020stability,Chang2021,Yin2022,Khansari2011,Khansari2014}. \cite{Chow2018, DBLP:journals/corr/abs-1901-10031} solves constrained MDPs with Lyapunov methods to constrain a policy search space during each training iteration. They formulate a constraint value function as a Lyapunov function and update the policy with Lyapunov constraints. The work of~\cite{DBLP:journals/corr/abs-2004-14288} constructs candidate Lyapunov functions from value functions in an actor-critic framework, using Lyapunov decreasing condition as critic value to enhance stability properties of neural control policies. The work of~\cite{Chang2021} performs self-learning of almost-Lyapunov functions, used as a critic function to accelerate policy training.

A Lyapunov-based approach for LfD problems by~\cite{Yin2022} relies on Lyapunov theory and local quadratic constraints to establish safety and stability guarantees for deep neural network control systems. The methods assume that the environment is a linear dynamical system and do not consider more complex environments. Other Lyapunov-based methods in~\cite{Khansari2014, Shavit2018, Figueroa2018} learn globally asymptotic system dynamics (transition model) and then plan for trajectories toward the goal state by leveraging the prediction models, with error compounding issues similar to BCO~\cite{BCO,Laskey2016,RossBagnell2010,Ross2011}.  

\section{Preliminaries}
\label{section:Prelim}

\noindent{\bf Markov Decision Processes and Learning from Observations.} We consider imitation learning in Markov Decision Processes (MDPs)~\cite{MarkovDecisionProcessBook, SuttonRL}. MDPs are defined as $\langle\mathcal{S},\mathcal{A},P,r,\gamma\rangle$, where $\mathcal{S}$ is the state space, and $\mathcal{A}$ is the action space. $P(s_{t+1} | s_t,a_t)$ is the transition probability of reaching state $s_{t+1}$ after action $a_t$ is taken at state $s_t$, where $t$ denotes a time-step but does not directly affect the transition probability. In the standard RL setting, the agent receives $r(s,a,s')$ where $r:\mathcal{S}\times\mathcal{A}\times\mathcal{S}\rightarrow \mathbb{R}$ is the reward function, and $\gamma$ is the discount factor. In the imitation learning setting, the reward function is assumed unknown, and the goal is to train the agent to perform the task given expert observation trajectories, so we can write the MDP for imitation problems as $\langle\mathcal{S},\mathcal{A},P,C\rangle$ where $C$ is the cost function that measures the deviation. In the LfO setting, there is no action information in the expert trajectories, so the cost function for the learner is defined as $C:\mathcal{S}\times\mathcal{S}\rightarrow \mathbb{R}$ which only assigns cost after comparing states between the learner and the expert. The LfO agent attempts to learn a policy $\pi_\phi:\mathcal{S}\times\mathcal{A} \rightarrow [0,1]$ such that sampling from policy $\pi_\phi$ produces a distribution of state-action-cost tuples $\{(s_i,a_i,c_i)\}$ under environment dynamics. The goal of LfO is to minimize cumulative of the cost function $C$ along trajectories generated by the agent's policy $\pi_\phi$.

%\todo[inline]{MDP before dynamical systems}

\noindent{\bf Stability and Lyapunov Methods.} In stabilization problems, the agents control dynamical systems: 
% agents are dynamical systems of the form
\begin{equation}
\label{eqn:dynamics}
    \dot x(t)= f(x(t),u(t)), u(t)=h(x(t)),x(0)=x_0,
\end{equation}
where $x(t)$ takes values in an $n$-dimensional state space $X\subseteq \mathbb{R}^n$, $f:X\rightarrow \mathbb{R}^n$ is a Lipschitz-continuous vector field, $h: X\rightarrow \mathbb{R}^m$ is a control function. Each $x(t)\in X$ is called a state vector and $u(t)\in \mathbb{R}^m$ is a control vector. The notion of stability is then formally defined as:
\begin{definition}[Lyapunov and Asymptotic Stability] A system of Eq.~\ref{eqn:dynamics} is \textit{Lyapunov stable} at the origin $x=0$, if for all $\epsilon >0$, there exists $\delta=\delta(\epsilon)>0$ such that for all $\|x(0)\|<\delta$, then $\|x(t)\|<\epsilon$ for all $t\geq0$. The system is \textit{locally asymptotically stable} at the origin if it is Lyapunov stable and there exists $\delta >0$ such that if $\|x(0)\|<\delta$, then $\lim_{t\rightarrow \infty}x(t)=0$. $\|\cdot\|$ is typically the Euclidean norm.
\end{definition}
\begin{definition}[Lie Derivatives] Consider the system in Eq.~\ref{eqn:dynamics} and let $V:X\rightarrow \mathbb{R}$ be a continuously differentiable function. The \textit{Lie derivative} of $V$ over $f$ is defined as
\begin{equation}
\label{eqn:liederivative}
    L_fV(x) = \sum\limits_{i=1}^n \frac{\partial V}{\partial x_i} \frac{\mathrm{d}x_i}{\mathrm{d}t} = \sum\limits_{i=1}^n \frac{\partial V}{\partial x_i} \dot x_i(t).
\end{equation}
The Lie derivative measures the change of $V$ over time along the direction of the system dynamics.
\end{definition}

\begin{definition}[Lyapunov Conditions for Asymptotic Stability]
\label{def:asymptotic}
Consider a controlled system Eq.~\ref{eqn:dynamics} with an equilibrium at the origin, i.e. $\exists u \in \mathbb{R}^m$ so $f(0,u)=0$. Suppose there is a continuously differentiable function $V:X\rightarrow \mathbb{R}$ satisfying $V(0)=0$; $\forall x \in X\setminus\{0\}$, $V(x)>0$; and $L_fV(x)<0$. Then $V$ is a \textit{Lyapunov function}. The system $f$ is asymptotically stable at the origin if Lyapunov function $V$ can be found.

\end{definition}
We train neural networks to learn approximate Lyapunov landscapes based on the expert state observations by enforcing Lyapunov conditions to be satisfied along expert trajectories. Because the procedure is learning-based and cannot guarantee the Lyapunov conditions throughout the entire state space, we use the name {\em Lyapunov-like proxy model} for our LfO setting. 

\noindent{\bf GAN-based Approaches in Imitation Learning.} Adversarial Imitation Learning approaches~\cite{GAIL, Fu2018, Henderson2018, Kostrikov2019, Sun2021} rely largely on the usage of Generative Adversarial Networks (GANs)~\cite{GAN}. The GAN architecture pits two neural networks against each other in order to make one neural network produce data distributions similar to that of the training data. The two neural networks are called the Generator ($G$) and Discriminator ($D$) respectively. $G$ attempts to fool $D$ by making its output distribution $p_g$ similar to training data distribution $p_{data}$ given data from a prior on input noise variables $p_z$. $D$ is trained to maximize the probability of assigning the correct label to both training data examples and samples from $G$, while $G$ is concurrently trained to minimize $\log(1 - D(G(z)))$. The minimax two-player game can thus be formulated with value function $W(D,G)$~\cite{GAN}:
\begin{multline}
\label{eqn:GANeq}
    \min_G\max_D W(D,G)= \\
    \E_{x\sim p_{data}}[\log D(x)] + \E_{z\sim p_z}[\log(1 - D(G(z)))]
\end{multline}
In GAIL~\cite{GAIL}, $p_{data}$ is the expert trajectory of state-action pairs, and $G$ is the imitation policy $\pi_\phi$ that needs to be trained and from which state-action pairs are sampled. The goal is to make the learner's state-action distribution be close to that of the expert. For LfO problems, GAIfO~\cite{GAIfO} uses similar paradigm, replacing state-action pairs with state transitions.

\section{Learning Stabilization Control}% from Observations}
\label{section:AlgExplanation}
%%
%% New Introduce Algorithm
%%
At a high-level, our algorithm comprises two steps. We first train a neural Lyapunov-like proxy model using the expert trajectories. This step of training uses violation of the Lyapunov conditions as the loss function, adapting the conditions for asymptotic stability from Definition \ref{def:asymptotic} to the LfO setting. 
%Learning the Lyapunov function from expert trajectories is important since if we intuitively guess Lyapunov function based on domain knowledge (such as squared error from equilibrium state), the learning rate can be slow, even slower than GAIfO.
The learned Lyapunov function subsequently provides the reward signal for training the learner's agent. For this second step of training to succeed, we need to define a transformation of the Lyapunov-like proxy model from the first step using convex functions. The overall algorithm called LSO-LLPM, short for Learning Stabilization Control from Observations through Learning Lyapunov-like proxy Models, is shown in Algorithm~\ref{alg:Algorithm}. Line $1-3$ is the first step of training the Lyapunov-like proxy model (Section \ref{section:learnLyap}), and Line $4-7$ is the second step of using Proximal Policy Optimization (PPO)~\cite{PPO} to optimize the learner's policy given a reward function derived from the Lyapunov-like proxy model (Section \ref{section:trainagent}).

\begin{algorithm}
\caption{LSO-LLPM}
\label{alg:Algorithm}
\begin{algorithmic}[1]
\Require Expert state-only trajectories: $\tau_E=\{(s,s')\}$, randomly initialized policy network $\pi_\phi$, randomly initialized Lyapunov-like proxy model $V_\theta$, and hyperparameters $c, \beta_1,\beta_2 \in\mathbb{R}^+$. 
\ForEach {$(s,s')\in\tau_E$}
\State Update $V_\theta$ by taking gradient descent steps on the following loss function:
\[ V_\theta^{2}(0) + \max(0,-V_\theta(s)) + \beta_1(c+(V_\theta(s')-V_\theta(s))/\Delta t)^2\]
\EndFor
\For {$i=0,1,2,...$}
\State Sample trajectories $\tau_i \sim \pi_{\phi}$
\State Update $\pi_\phi$ by performing PPO update steps with the following reward function:
\[g(V_\theta(s)) + \beta_2 \min(0, (V_\theta(s)-V_\theta(s'))/\Delta t)\]
\EndFor
\end{algorithmic}
\end{algorithm}

\subsection{Learning Neural Lyapunov-like proxy Models}
\label{section:learnLyap}

In the first step, we learn Lyapunov-like proxy models to capture the stabilization behavior of the expert. The shape of the region of attraction~\cite{lyapunovbook} and the convergence rate of this approximate Lyapunov landscape are important because we need to avoid misleading the learner in the second training step to pursue behaviors that cannot be achieved because of constraints in the underlying dynamics. For instance, when controlling a vehicle to track a given path, if we define an aggressive Lyapunov landscape to aim for a fast convergence rate, then the vehicle may be forced to attain a high speed and fail to stabilize. On the other hand, if the Lyapunov landscape is too smooth, then the learner may attempt to take conservative actions that are not sufficient to drive the system to the target equilibrium. 

We represent the Lyapunov-like proxy model over state space, $V_{\theta}:S\rightarrow \mathbb{R}$, as a neural network (typically two-layer fully connected networks are sufficient in our examples), which is randomly initialized. We sample observations from the expert trajectories and perform stochastic gradient descent on the parameters to minimize the violations of Lyapunov conditions using the following loss function:
\begin{equation}
\label{eqn:lyaploss}
    V_\theta(0)^2 + \max(0,-V_\theta(s)) + \beta_1(c + L_fV_\theta(s))^2
\end{equation}
Recall that $L_fV_{\theta}$ is the Lie derivative of $V_{\theta}$ along the system dynamics $f$ that we do not have knowledge of. Instead, we approximate the Lie derivative by the finite difference of the Lyapunov function over each discrete time step, $L_fV_\theta = {(V_\theta(s') - V_\theta(s))}/{\Delta t}$, where $s$ and $s'$ are consecutive state observations in the trajectory. We know that this is a good approximation when $\Delta t$ is small. 

The first term $V_\theta(0)^2$ ensures that the equilibrium point corresponds to a Lyapunov value of zero, which is the lowest across the state space because the second term $\max(0,-V_\theta(s))$ requires that the Lyapunov function value be non-negative at all sampled points, which ideally generalizes to other regions in the state space. The third term $\beta_1(c + L_fV_\theta(s))^2$ is a critical design. It controls the convergence rate as measured by the Lie derivative $L_fV_\theta$. Here we deviate from the standard Lyapunov conditions of only requiring the Lie derivative to be negative by forcing it to take the value of some positive constant rate $c$. With this requirement, the overall landscape $V_\theta$ will be shaped through learning so each step taken by the expert will be considered as a unit step toward stabilization. In this way, we capture the convergence rate by the Lyapunov-like proxy model which can then properly guide the learner in the second training step. We stop gradients for $V_\theta(s')$ in calculating $L_fV_\theta$.

\subsection{Policy Learning from the Lyapunov-like proxy Model}
\label{section:trainagent}
After learning the Lyapunov-like proxy model, we transform it into a reward function that the learner can maximize using standard policy optimization procedures such as PPO~\cite{PPO}. The reward is defined as:
\begin{equation}
\label{eqn:lyaptrain}
    g(V_\theta(s)) + \beta_2 \min(0, (V_\theta(s)-V_\theta(s'))/\Delta t)
\end{equation}
where $g(x)$ is a convex function (fixed through all environments) for scaling the values of the Lyapunov-like proxy models so that the learner can receive sufficiently strong reward feedback in each step. We also want to maintain stability when the agent is already close to the target equilibrium point. 
%We want to increasingly reward the policy as it gets closer to its Lyapunov equilibrium point $V_\theta(0)$. 
%This $g$ function transforms the Lyapunov function so that it becomes more desirable for the agent to be closer to the equilibrium state. 
Suitable choices of $g(x)$ include $-\log(x)$ and $-\log(1-e^{-kx^2})$ for some $k>0$. The second term of Eq.~\ref{eqn:lyaptrain} reduces the reward when the Lie derivative becomes positive, which prevents the learner from taking large steps near the equilibrium state. In this manner, we observe fast converge of the agent to the equilibrium state. Overall, the Lyapunov candidate function acts as a proxy for the reward function so that PPO will take steps to increase the reward and attempt to reproduce stabilization control policies that converge at a similar rate as the trajectories from the expert. 

\section{Experiments in Simulated Environments}
\label{section:Experiments}
We evaluate our algorithm LSO-LLPM for nonlinear control problems and compare with the state-of-the-art algorithm GAIfO. For fair comparison, we implement GAIfO using PPO without entropy loss. Evaluation environments include Automobile path-tracking control~\cite{Snider2009} and Acrobot~\cite{AcrobotAlborz}, as well as high dimensional tasks like Quadrotor~\cite{Rubi2020}, and Hopper Standing, and Walker Standing. Acrobot is a classical control task simulated within OpenAI Gym~\cite{OpenAI}, Automobile path-tracking control and Quadrotor were simulated using PythonRobotics~\cite{PythonRobotics}, and Hopper and Walker Standing environments were simulated with MuJoCo~\cite{MuJoCo}. We used NNs with $2-3$ hidden layers with $64 - 2048$ neurons.

\noindent{\bf Baselines and Evaluation Metrics.} We focus on comparing with the state-of-the-art on-policy LfO approach GAIfO~\cite{GAIfO} because of the same assumptions of having access to observations only and on-policy training of the learner without modeling the environment transitions. In particular, GAIfO has been shown to perform consistently better than BCO~\cite{BCO}. There are a range of other imitation learning baselines with different additional assumptions, like LfD methods with access to actions (GAIL~\cite{GAIL}, DAC~\cite{DAC}), and off-policy LfO methods (OPOLO~\cite{OPOLO}) that assume ability to pre-train inverse transition models to accelerate learning progress. This performance gain is orthogonal to on-policy learning objective in our setting. Consequently, we focus on comparing with GAIfO in the evaluation and analysis.

We evaluate the performance along the following metrics: First, we measure the overall performance of the learner with respect to a varying number of sampled expert trajectories that are provided to both algorithms. Second, we analyze the learning efficiency by the learning curves over time.

\begin{figure}[ht]
\centering
\includegraphics[width=\linewidth]{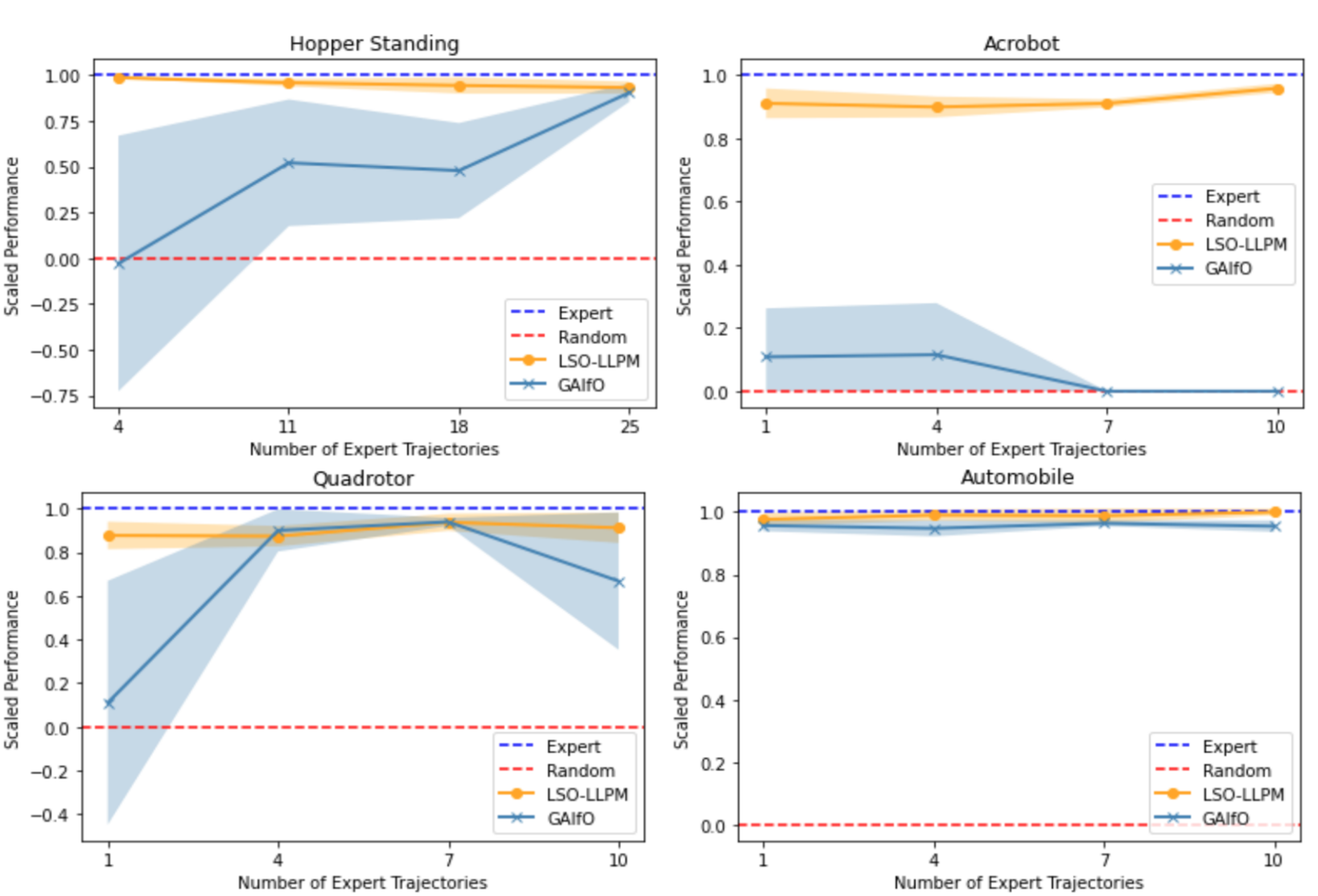}
\caption{Performance of learned policies with varying numbers of expert trajectories. The performance is normalized to be between 0 (average reward of random policy) and 1 (average reward of expert policy). The shaded areas show variance over 5 random seeds. We observe the proposed methods perform much better especially in environments that are harder to control, like Hopper % and Walker 
Standing, Acrobot, and Quadrotor.}
\label{fig:perftraj}
\end{figure}

\noindent{\bf Environments.} The nonlinear control tasks in each environment are specified as follows:

\noindent{\em - Acrobot:} The Acrobot environment consists of two links and two joints. In the initial state, both links are hanging down. The goal is to swing the links up so the tip of the link farthest from the pivot reaches the threshold in the shortest time. The state consists of information on the angles from the joints as well as their angular velocity, and the agent can actuate the joint between the links.

\noindent{\em - Automobile path-tracking control:} Autonomous driving is a control problem in which using speed commands the agent needs to follow a target path. In the environment, the state space is four dimensions, namely the difference between target speed and vehicle speed $V_t-v(t)$, the angular error $\theta_e(t)$, the distance to the path $d_e(t)$, and vehicle speed $v(t)$. The action space has two dimensions, namely the acceleration $a(t)$ and a steering control $\delta(t)$.

\noindent{\em - Quadrotor control:} We also test our algorithm in the $6$-degree-of-freedom Quadrotor model. This has four control inputs and twelve state variables. The control inputs $\Omega_1$, $\Omega_2$, $\Omega_3$, and $\Omega_4$ are the angular velocities of each rotor. The state variables are the inertia frame positions $(x,y,z)$, velocities $(\dot x,\dot y,\dot z)$, rotation angles $(\phi, \theta, \psi)$, and angular velocities $(\dot\phi, \dot\theta, \dot\psi)$. More details regarding the implementation and dynamics of the Quadrotor can be found in~\cite{Rubi2020, PythonRobotics}. This control problem is known to be hard for policy gradient methods but is solved with learning from demonstrations.

\noindent{\em - Hopper and Walker Standing:} We use MuJoCo Hopper and Walker and change task reward to formulate the stabilization control problem of standing in upright position. In Hopper Standing, there is one leg with $3$ joints. In Walker Standing, there are two legs with $6$ joints. In both environments, the agent's goal is to maintain standing state without falling down (losing balance) for the longest time. We show graph results for Hopper Standing as the results are similar.

\begin{figure}[ht!]
\centering
\includegraphics[width=\linewidth]{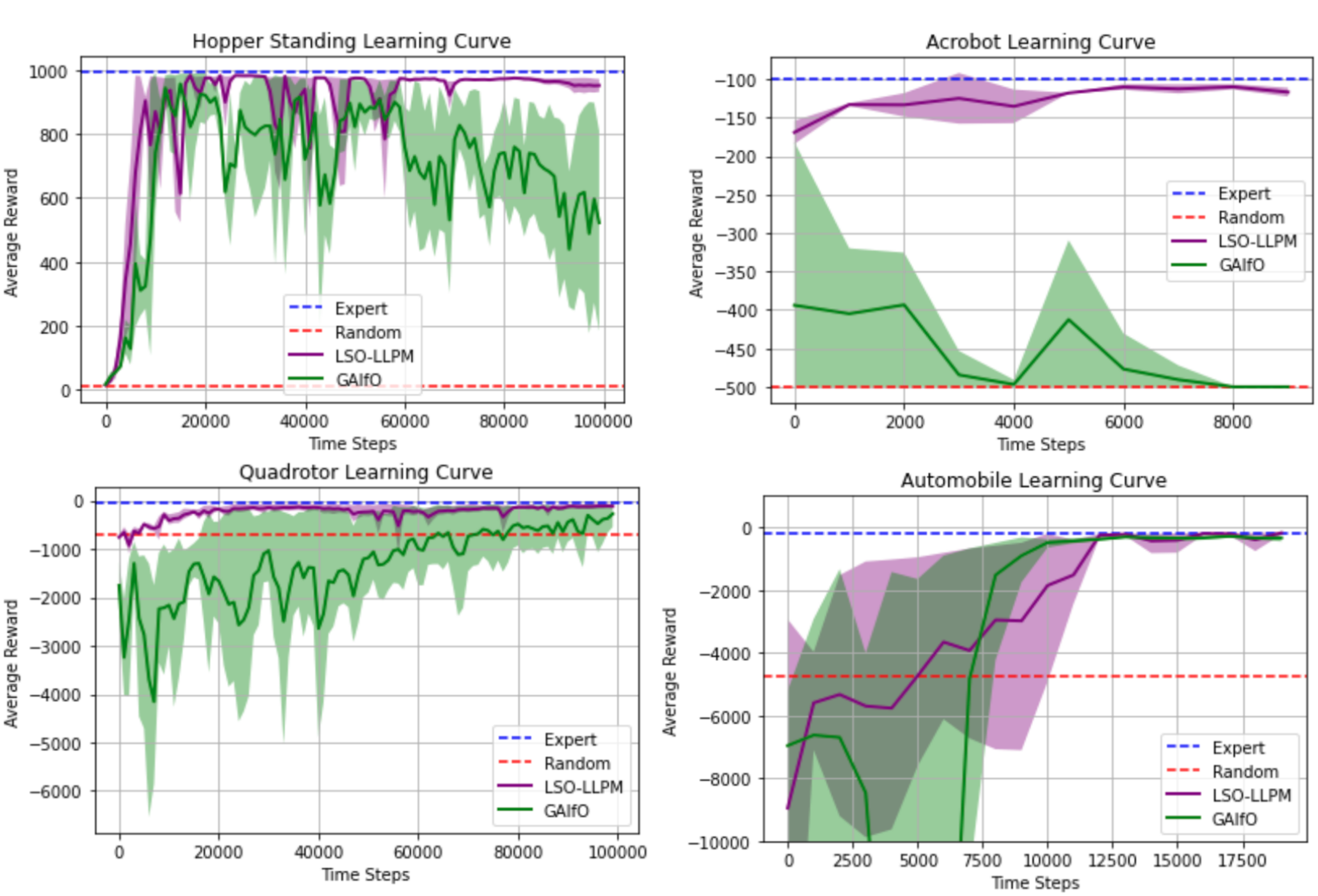}
\caption{Comparing learning curves using $10-18$ expert trajectories (fixed number across different methods in each environment) over $5$ random seeds. Environments in first row show distribution matching approaches in GAN-based methods often experience difficulty in making consistent learning progress while Lyapunov-like proxy models generate landscapes that are suitable for stabilization tasks and thus achieve stable learning performance.}

\label{fig:learningcurve}
\end{figure}

\subsection{Overall Performance}

\noindent{\bf Results: LfO Performance.}
We examine the performance of the policies trained by LSO-LLPM and GAIfO with different number of expert trajectories in Figure~\ref{fig:perftraj}. LSO-LLPM reaches at least $85\%$ of the expert performance in all environments 
%(except Walker Standing) 
for all number of expert trajectories tested. As shown in the figure, LSO-LLPM consistently performs better than GAIfO for all environments and number of expert trajectories. In particular, in Hopper Standing 
%, Walker Standing, 
and Acrobot, LSO-LLPM is able to perform well, even when GAIfO performs not much better than the random policy baseline. This difference illustrates the important use of Lyapunov-like proxy models that capture challenging control problems in these environments due to underlying nonlinear dynamics. In these cases, GAN-based approaches performing distribution matching are often not sufficient for finding good control policies for stabilization.

\begin{figure*}[t]
\centering
\includegraphics[width=1\textwidth]{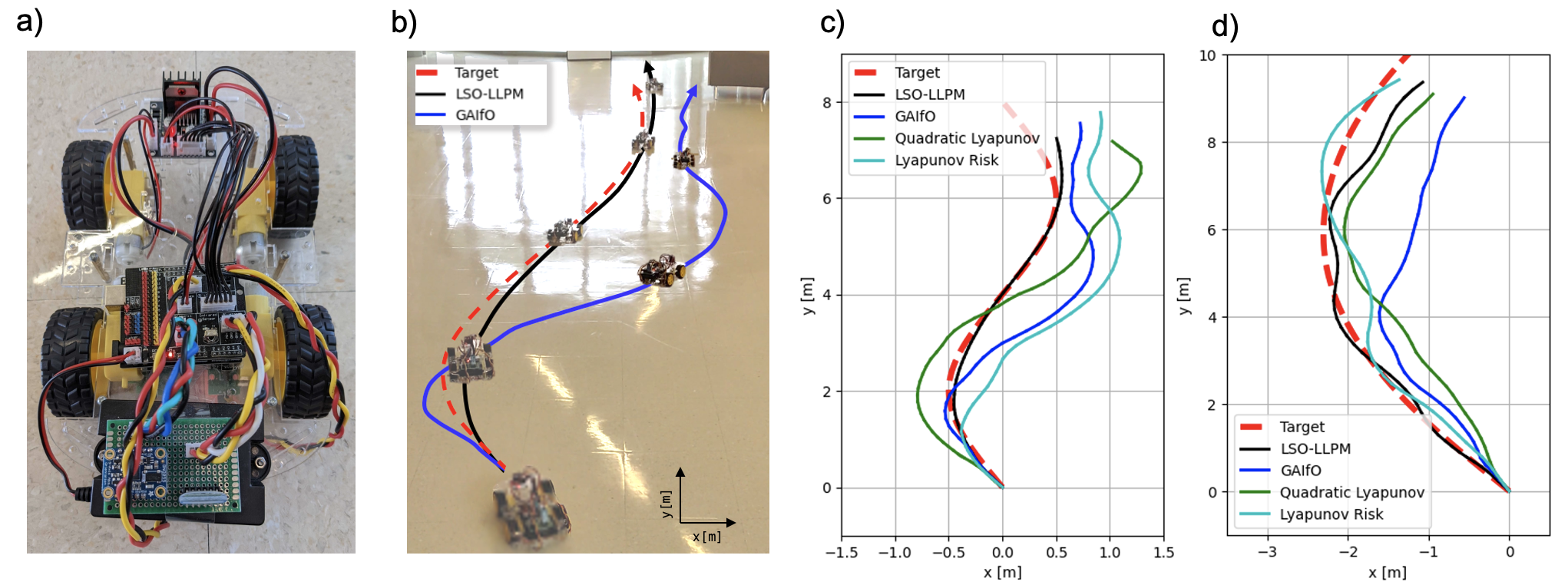}
\caption{(a) Car robot used for the hardware experiments. (b) Tracking target path I. The car controlled by LSO-LLPM (black line) tracked the target path (red dash line) better than GAIfO (blue line). (c), (d) All trajectories driven by LSO-LLPM, GAIfO, Quadratic Lyapunov, and Lyapunov Risk. Our proposed method tracked both target paths fairly well compared with the other methods. (c) shows the result of tracking target path I and (d) shows the result of tracking target path II.}\label{fig:car_experiment}
\end{figure*}

\noindent{\bf Results: Learning Efficiency.} The learning curves of LSO-LLPM and GAIfO in each environment are shown in Figure~\ref{fig:learningcurve}. In each environment, we use the number of expert trajectories (ranging between $10$ to $18$) that corresponds to the high performance cases in Figure~\ref{fig:perftraj}. We see that across all environments, the LSO-LLPM methods can generally achieve fast learning progress with small variance across different random seeds. In comparison, GAIfO can learn well in simple control tasks such as simulated Automobile, but struggles to make progress in harder environments such as Hopper Standing
%, Walker Standing,
and Acrobot. Note that this difficulty is not only determined by the dimensionality of the action space but also by how  difficult it is to find control policies that can allow the learner to imitate the expert trajectories, and the LfO setting further increases that difficulty. For instance, although Acrobot is low-dimensional, the stabilization control problem is still very challenging. Overall, we observe Lyapunov-like proxy models can generate landscapes that are particularly suitable for stabilization tasks for efficient training of the learner in hard control tasks.

\section{Experiments in Hardware Environments}
\label{section:ExperimentsHW}
We present the evaluation of performance of automobile path-tracking control driven by LSO-LLPM, GAIfO, Quadratic Lyapunov, and Lyapunov Risk.

\noindent\textbf{Quadratic Lyapunov} In Lyapunov stability theory, a typical choice is the quadratic form (sum of squared differences between state features and their ideal locations/orientations) as default Lyapunov landscape. This is provably optimal for linear systems and often works well for mildly nonlinear systems. We consider whether a quadratic model can guide the learner without learning a Lyapunov-like proxy model.

\noindent\textbf{Lyapunov Risk} We also consider learning the Lyapunov Function with the generic Lyapunov Risk Loss as proposed in~\cite{NLC} instead of LSO-LLPM's loss expression in Eq.~\ref{eqn:lyaploss}. This would demonstrate the importance of enforcing constant Lie derivative on expert trajectories in guiding the learner. 
\noindent\textbf{Hardware settings}
We tested the control task using the car robot shown in Fig. \ref{fig:car_experiment} (a). One IMU sensor and two photoelectric encoders are on it to calculate its velocity $v(t)$ and orientation $\theta(t)$ each time step. With these values, we obtain the difference between target speed and vehicle speed $V_t - v(t)$, the angular error $\theta_e(t)$, and the distance to the path $d_e(t)$, which are used as each policy's input. Given these input values, the policies determine acceleration $a(t)$ and steering control $\delta(t)$ at the next time step. In this hardware experiment, we set $\Delta t = 0.5 \;$[s], target speed $V_t = 0.3\;$[m/s]. The policies are trained under a simulator with these parameter settings, and they are directly deployed to the robot car for testing.

\noindent\textbf{Results}
We tested the performance of automobile path-tracking with two different target paths, Target path I and II. Fig. \ref{fig:car_experiment} (b) and (c) are the results of tracking target path I, and Fig. \ref{fig:car_experiment} (d) shows the result of tracking target path II. The red dot lines are the target paths given in advance, and the closer the trajectories are to them, the better their control policies. As seen from these figures, the car driven by LSO-LLPM (black lines) tracked the target paths fairly well compared with the other methods. In particular, GAIfO (blue lines) showed low path tracking performance as the trajectories controlled by it gradually deviated from the target paths. The main strength of our approach is obtaining stable control policies through LfO training process by utilizing Lyapunov-like proxy models. This enables more stable tracking performances even for the hardware experiments, in which are various disturbances such as sensor noises and discrepancies between simulator and real hardware modeling.

\section{Discussion and Conclusion}
%\section{Discussion and Conclusion}

%For nonlinear autonomous (time invariant dynamics) control systems, for our Lyapunov based algorithm, for both low and high dimensional environments. In this work, 
We have introduced a novel model-free Lyapunov-based method to accelerate Learning from Observations for stabilization control problems by introducing intermediate proxy models between the expert and the agent policy based on Lyapunov stability theory. Our LfO training process first learns a Lyapunov landscape model from the expert state sequences and then transforms the learned model to guide the training of the learner's policy. We showed the proposed methods can capture stabilization control behaviors that take into account underlying dynamics so the learner's agent can successfully recover stable control policies through policy optimization. We evaluated the proposed methods in various real and simulated nonlinear stabilization control environments and observed better learning efficiency compared to the state-of-the-art GAN-based approaches. 

We primarily focused on stabilization to a fixed equilibrium. To handle more general control tasks, it is possible to extend our approach by incorporating time in the Lyapunov-like proxy models. Such extension may work on more control environments with general locomotion tasks.

\newpage
\bibliographystyle{unsrt}
\bibliography{ref}

\end{document}